\documentclass[a4paper,fleqn]{cas-dc}

\usepackage[numbers]{natbib}
\def\tsc#1{\csdef{#1}{\textsc{\lowercase{#1}}\xspace}}
\tsc{WGM}
\tsc{QE}
\tsc{EP}
\tsc{PMS}
\tsc{BEC}
\tsc{DE}
\usepackage{graphicx}
\graphicspath{ {./images/} }
\usepackage{subfig}
\usepackage{hyperref}

\begin{document}
\let\WriteBookmarks\relax
\def\floatpagepagefraction{1}
\def\textpagefraction{.001}

\shorttitle{TransCORALNet}

\title [mode = title]{TransCORALNet: A Two-Stream Transformer CORAL Networks for Supply Chain Credit Assessment Cold Start}                      

%\tnotetext[1]{This document is the results of the research
   %project funded by the National Science Foundation.}

\let\printorcid\relax
\author{Jie Shi}
\author{Arno P. J. M. Siebes}
% Corresponding Author
\author
{Siamak Mehrkanoon}
\cormark[1]

\cormark[1]

%Footnote of the first author

%\ead{j.shi1@uu.nl}

%\ead[url]{www.cvr.cc, cvr@sayahna.org}

\credit{Conceptualization of this study, Methodology, Software}

\affiliation{organization={Department of Information and Computing Sciences, Utrecht University},
    city={Utrecht},
    % citysep={}, % Uncomment if no comma needed between city and postcode
    %postcode={1043 NX}, 
    country={The Netherlands}}

\cortext[cor1]{Corresponding author: Siamak Mehrkanoon, email address: s.mehrkanoon@uu.nl}

\fntext[Github]{https://github.com/JieJieNiu/TransCORALNet}

\begin{abstract}
This paper proposes an interpretable two-stream transformer CORAL networks (TransCORALNet) for supply chain credit assessment under the segment industry and cold start problem. The model aims to provide accurate credit assessment prediction for new supply chain borrowers with limited historical data. Here, the two-stream domain adaptation architecture with correlation alignment (CORAL) loss is used as a core model and is equipped with transformer, which provides insights about the learned features and allow efficient parallelization during training. Thanks to the domain adaptation capability of the proposed model, the domain shift between the source and target domain is minimized. Therefore, the model exhibits good generalization where the source and target do not follow the same distribution, and a limited amount of target labeled instances exist. Furthermore, we employ Local Interpretable Model-agnostic Explanations (LIME) to provide more insight into the model prediction and identify the key features contributing to supply chain credit assessment decisions. The proposed model addresses four significant supply chain credit assessment challenges: domain shift, cold start, imbalanced-class and interpretability. Experimental results on a real-world data set demonstrate the superiority of TransCORALNet over a number of state-of-the-art baselines in terms of accuracy. The code is available on GitHub \footnote{Github}.
\end{abstract}

% \includegraphics{figs/grabs.pdf}

%\item Research highlights item 1
%\item Research highlights item 2
%\item Research highlights item 3

% Keywords

\maketitle

\section{Introduction}

Machine learning can potentially reduce banks' credit losses by $20\%$ to $40\%$, which can lead to revenue increase by $5\%$ to $15\%$, and cut credit decision time by $20\%$ to $40\%$, according to a 2021 report from McKinsey \cite{mckinsey}. Conventional machine learning (ML) has benefited from financial big data and assumes that the data follow the same distribution. Applying a range of ML methods in personal credit has been widely studied over the past few years. However, unlike personal credit, very few studies address ML algorithms to supply chain credit assessment \cite{yang2022supply}. Many financial institutions still assess supply chain credit through manual inspections, and current assessment approaches require appropriate industrial knowledge. Compared to machine learning models, the statistic models are easier to be explained, which is crucial in credit assessment decision-making. Therefore, statistical rule-based models have been extensively applied in credit risk assessment \cite{kozodoi2022fairness}.

The application of machine learning in supply chain credit faces the difficulties of domain shift, cold start, imbalanced classes, and interpretability. Conventional supervised machine learning is known to be limited by three facts: distribution mismatch between source and target, insufficiently labeled instances, and imbalanced classes. However, supply chain credits are usually categorized into different types by the segment industries or the circular supply chain (e.g., retail, logistics, manufacturing, construction, medical and internet industries). The behaviors of borrowers from different circular supply chain are commonly affected by the diverse and sophisticated industrial macro environment, industrial patterns, product and service forms, business modes, and business organization forms. As a consequence, the data of different borrower groups may follow different distributions. In addition, the number of supply chain borrowers could be more inadequate than the number of personal borrowers. The sample data of the core enterprise and its suppliers and buyers in specific supply chain circles are usually insufficient at the early stage. Without plenty historical data, financial institutions will have little knowledge of their new borrowers' behavior and histories. The cold start problem arises when there is not enough data to train a reliable model. At the same time, the class imbalanced problem is common as the number of defaulters is usually much less than the number of non-defaulters. 
Moreover, financial institutions prefer to apply explainable AI (XAI) models in credit assessment, which help them meet financial regulatory requirements. Explainable Artificial Intelligence (XAI) models enable financial institutions to offer transparent and comprehensible explanations for their credit decisions. This is crucial for building trust among borrowers and for avoiding potential black-box legal issues.

In recent years, transformer based models have become one of the most popular and practical approaches in many Artificial Intelligence tasks. One of the most important component in transformer architecture \cite{vaswani2017attention} is self-attention mechanism, which allows the model to focus on different parts of the input sequence and generate context-aware representations. Domain adaptation is a rapidly growing research area in machine learning that addresses the challenge of transfer learning from a source domain to a target domain where the distribution differs \cite{mehrkanoon2017regularized,csurka2017domain,mehrkanoon2019cross,li2020model,gal2022stylegan}. Recent studies have shown deep learning-based domain adaptations, such as discrepancy-based and adversarial-based domain adaptations, have achieved remarkable performance in various tasks. For instance, Correlation Alignment for Deep Domain Adaptation (Deep CORAL) \cite{sun2016deep} which belongs to discrepancy-based domain adaptation has successfully been applied in some fields, such as health and medical \cite{zhong2023deep}, mechanical intelligent diagnosis \cite{jiao2019classifier}, and financial fraud detection \cite{lebichot2020deep}. Its CORAL loss function aligns the second-order statistics of the source and target distributions with a linear transformation. While the application of transformer and domain adaptation in NLP and computer vision problems is largely unexplored. However, not much research has explored their use in supply chain credit assessment. This paper proposes the two-Stream transformer CORAL networks for supply chain credit assessment. To the best of our knowledge, this is the first attempt to adapt and apply TransCORALNet in supply chain credit assessment. Our work leads to five distinct novel contributions:
\begin{itemize} 
\item To solve the domain shift problem, we employ a CORAL loss function. This function can effectively minimize and adapt the domain shift from established to emerging industries.
\item To facilitate the cold start problem, we adopt a GAN-based method (CTGAN) \cite{xu2019ctgan} to model tabular data distribution. It can generates synthetic data by learning the probability distribution of the features in the target domain. These synthetic data compensate for insufficient samples in the target domain. Moreover, abundant synthetic data helps the network learn distributional differences.
\item To resolve the label imbalance problem, we apply a weighted cross entropy loss function, which gives the defaulting samples a relatively high weight. It can significantly improve the recall and F1 scores of the minority class.
\item To address the black-box problem, we provide feature-to-feature interpretability through the attention mechanism and local feature-to-output explainability through Local Interpretable Model Agnostic Interpretation (LIME) \cite{ribeiro2016lime}. The self-attention mechanism can extract the relationship among features. In contrast, the LIME can efficiently maps the features to the output classification.
\end{itemize}

This paper is organized as follows. A brief overview of related research on supply chain credit assessment using machine learning approaches is presented in Section \ref{section:sec2}. In Section \ref{section:sec3}, we describe the proposed TransCORALNet architecture. Section \ref{section:sec4} focuses on the conducted experiments and the obtained results. A discussion of the results is given in Section \ref{section:sec5}. A conclusion and implications for future research is drawn in Section \ref{section:sec6}.

\begin{figure*}[htp]
\centering
\includegraphics[width=16cm]{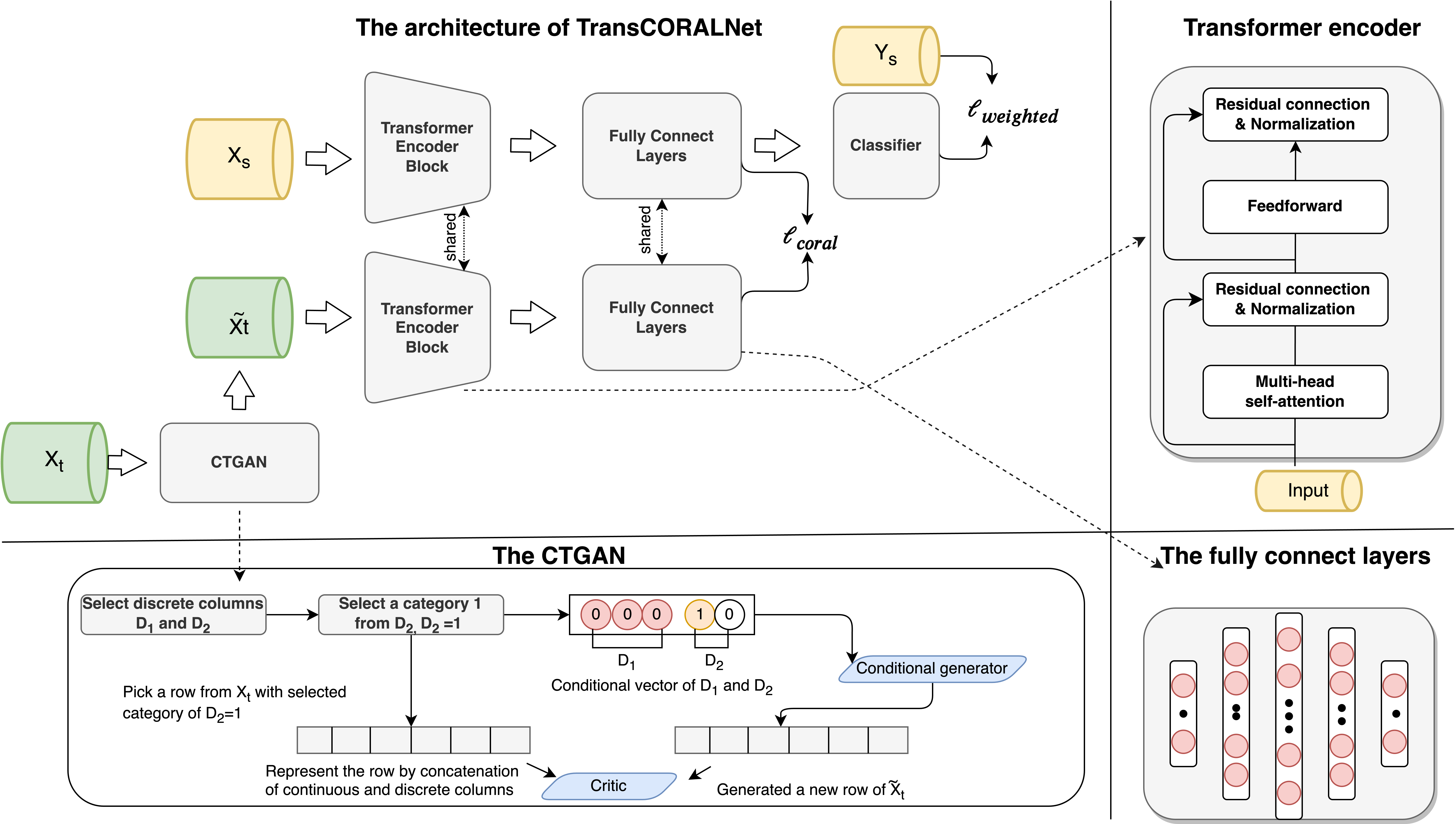}
\caption{The architecture of TransCORALNet. This graph depicts the overall framework of our proposed model, the components of the transformer encoder block, the fully connect layers, and the CTGAN procedures.}\label{fig1}
\end{figure*}

\section{Related work}\label{section:sec2}
\subsection{Domain adaptation}
Over the last few years, many different Domain Adaptation (DA) methods have been studied that address the domain shift problem. DA can be categorized into discrepancy-based, adversarial-based, multi-domain, hybrid, and emerging DA approaches \cite{singhal2023domain}. Discrepancy-based DA models aim to reduce domain shift by fine-tuning the deep network. It uses an extension of shallow learning with domain-specific loss function, such as maximum mean discrepancy (MMD) \cite{tzeng2014deep}, Deep CORAL \cite{sun2016deep}, contrastive domain discrepancy \cite{kang2019contrastive} and Kullback-Leibler (KL) divergence \cite{zhuangsupervised}. On the other hand, adversarial-based DA models draw on the core ideas of generative adversarial networks (GAN) \cite{goodfellow2020generative}. The discriminator in adversarial domain adaptation should be able to identify the domain distribution of source and target domains. The Domain Adversarial Neural Network (DANN) \cite{ganin2016domain} is a typical adversarial domain adaptation. Its Siamese network consists of two parallel streams, one processing the source data and the other processing the target data. The model parameters are optimized to minimize the loss of the label classifier, which classifies classes, and maximize the loss of the domain classifier, which discriminates between source and target domains. Using a similar idea to DANN, Long et al. \cite{long2018conditional} presented conditional adversarial domain adaptation (CADA), a framework that conditions the adversarial adaptation models on discriminative information conveyed in the classifier predictions. 

\subsection{DA in personal credit assessment}
Several studies on DA, primarily aimed at personal credit, have been conducted. Allowing for training on a source domain with abundant credit samples and subsequent application to a target domain with limited data, Huang and Chen \cite{huang2018domain} employed domain adaptation in P2P credit assessment using 434,419 samples. Chen et al. \cite{chen2020domain} proposed an improved DA based on structural similarity weighted average discrepancy loss function and applied it to 0.35 billion personal credit samples. This DA loss function can determine the contribution of each sample to the integral measure of the performance on the target domain. On the fraud detection problem, Lebichot et al. \cite{Lebichot} presented and discussed several self-supervised and semi-supervised domain adaptation classifiers based on 200 million transaction data. To address the issue of limited small business credit data, Suryanto et al. \cite{suryanto2022credit} applied a domain adaptation approach to transfer 100,000 personal credit data records to a target dataset consisting of 13,794 records of small businesses.

\subsection{Self-attention mechanism}

The core of the transformer \cite{vaswani2017attention} is the self-attention mechanism, which has become one of the most essential concepts in deep learning. It is inspired by human biological systems, which selectively focus on specific elements while filtering out the less relevant ones. Self-attention is a category of attention mechanisms that allows each element in a sequence to attend to all other elements, capturing long-range dependencies and relationships effectively \cite{vaswani2017attention}. Self-attention has been successfully 
used in various tasks, including NLP \cite{hu2020introductory}, computer vision \cite{ramachandran2019stand}, medical \cite{abdellaoui2021enhancing} and weather nowcasting \cite{yang2022aa}. In recent years, there has been an increasing amount of literature on the interpretability of self-attention. Although Jain and Wallace \cite{jain2019attention} designed experiments to find examples supporting that standard attention modules do not provide meaningful explanations, other researchers have emphasized that the attention scores can capture the relationship between the input and the output. For instance, many of the assumptions made in the research  \cite{jain2019attention} by Jain and Wallace are challenged by Wiegreffe and Pinter \cite{wiegreffe2019attention}, which indicated that the prior work \cite{jain2019attention} does not disprove the usefulness of attention mechanisms for explainability; Won et al.\cite{won2019toward} focused on visualizing and understanding the proposed self-attention based music tagging model; Akula and Garibay \cite{akula2021interpretable} exemplified the interpretability of the self-attention mechanism, showing that multi-head self-attention can effectively interpret the sarcastic cues in the input text; Rajabi and Garibay \cite{renault2023sar} introduced an explainable Small Attention Residual UNet for precipitation nowcasting tasks; A recent study by Zhao et al. \cite{bhan2023evaluating} assessed how attention coefficients can help in providing interpretability. 

\subsection{Oversampling of tabular data}

Generative Adversarial Networks (GAN) is a promising technique for addressing insufficient data and class imbalance problems. GAN offers a unique solution to this problem by generating synthetic samples. To date, many studies have adopted GAN to generate synthetic data. Oh et al. \cite{oh2019oversampling} presented an oversampling method using an outlier detectable generative adversarial network (OD-GAN) to solve the imbalance class problem. Another recent study by Zheng et al. \cite{zheng2020conditional} adopt the conditional generative adversarial network-gradient penalty-based approach to overcome the imbalanced data classification problem. In the context of credit risk assessment, one often encounters tabular data. Tabular data is one of the various data types for machine learning. Many researches on oversampling tabular data have been conducted to model diverse data types, including continuous and categorical variables. For example, Xu et al. \cite{xu2019ctgan} proposed a generative adversarial network (CTGAN) with Bayesian methods for tabular data synthesis; Similarly, Zhao et al. \cite{zhao2022ctab} introduced CTAB-GAN, a novel conditional tabular GAN architecture with encoders targeting mixed continuous-categorical variables and variables with unbalanced or skewed data; Rajabi and Garibay \cite{rajabi2022tabfairgan} developed a Generative Adversarial Network for tabular data generation; Zhao et al. \cite{zhao2022fct} presented a Fourier Conditional Tabular Generative Adversarial Network (FCT-GAN) on high dimensional tabular dataset.

\subsection{Explainable AI in credit assessment}

Explainable Artificial Intelligence (XAI) focuses on making AI systems transparent and interpretable \cite{gunning2019xai}. It addresses the {\lq black box\rq} problem by providing insights into how AI models make decisions. Surveys such as that conducted by Došilović et al. \cite{8400040} described the challenges of applying XAI solutions from a business perspective; Kruse et al. \cite{kruse2019artificial} analyzed several drivers and inhibitors of AI application in the finance industry. The results highlighted that explainability has become the challenge of AI adoption in this industry. A survey by Kuiper et al. \cite{kuiper2022exploring} described the regulators' and regulated entities' perceptions of XAI applications in the financial sector. Bussmann et al. \cite{bussmann2020explainable} proposed an interpretable AI model that can be used for Fin-tech risk management, particularly for measuring the risks arising when lending using peer-to-peer lending platforms; Misheva et al. \cite{misheva2021explainable} implemented two advanced model agnostic interpretation techniques for ML credit scoring models, namely Local Interpretable Model Agnostic Interpretation (LIME) and Shapley Additive exPlanations (SHAP), apply to the P2P Lending dataset; Gramegna and Giudici \cite{gramegna2021shap} evaluated two XAI models, LIME and SHAP, on real Small and Medium Enterprises data; Fritz-Morgenthal et al. \cite{fritz2022financial} aimed to provide practical advice on establishing a risk-based governance and testing framework and discusses using state-of-the-art technologies, methods, and platforms to support XAI.

\section{Methodology}\label{section:sec3}
The historical data of established supply chain borrowers is used as the source domain, while the samples of newly acquired borrowers form the target domain. The source domain features and labels are denoted by $X_{s}=\left\{x^{(i)}_{s}\right\}_{i=1}^{M}$, with $x_{s} \in \mathbb{R}^{d}$, and $Y_{s}=\left\{y^{(i)}_{s}\right\}_{i=1}^{M}$ respectively, whereas the target domain features and labels are denoted by $X_{t}=\left\{x^{(i)}_{t}\right\}_{i=1}^{N}$, with $x_{t} \in \mathbb{R}^{d}$, and $Y_{t}=\left\{y^{(i)}_{t}\right\}_{i=1}^{N}$. Here, the source and target domain instances, have the same dimension $d$. However, the number of instances in source domain is significantly greater than that of target domain, i.e., $M$ $\gg$ $N$. We have conducted a multivaritate Kolmogorov-Smirnov test \cite{justel1997multivariate} and the results show that the  marginal probability distribution of the source data ($P_{x_s}$) is different from that of the target data ($P_{x_t}$), yielding the domain shift problem \cite{sun2011two}. Hence, the domain adaption based techniques can be applied. Domain adaptation can benefit from a large amount of target training data \cite{ganin2015unsupervised}.
Therefore, we generate synthetic target training instances with the same quantity as the source domain, which are denoted by $\tilde{X}_{t}=\left\{\Tilde{x}^{(i)}_{t}\right\}_{i=1}^{M}$, with $\Tilde{x}^{(i)}_{t} \in \mathbb{R}^{d}$.

\subsection{Overall architecture}

Our proposed model builds upon and extends a two-stream CORAL architecture. The diagram that summarizes our proposed two-stream transformer CORAL networks (TransCORALNet) is depicted in Fig.\ref{fig1}. As opposed to \cite{sun2016deep}, here the encoder transformer is used to obtain new embeddings of the source and target data, leveraging its self-attention mechanism to capture dependencies within the input, thereby enhancing the model's ability to represent and interpret features for improved task performance. The TransCORALNet architecture (see the upper left of Fig.\ref{fig1}) consists of transformer encoder blocks, fully connected layers, and a final classifier layer. The upper stream receives the source data while the input of the lower stream is synthetic target training data which are generated by the CTGAN approach \cite{xu2019ctgan}. Here the encoder blocks and fully connected layers of two streams have shared weights. The CORAL loss \cite{sun2016deep} is used after the fully connected layers to minimize the domain shift. In addition, the source domain classification applies the weighted cross-entropy loss after the final classifier. The subsequent sections will provide a more detailed explanation for each component of the architecture.

\subsubsection{Synthetic target training}

A larger target domain dataset provides more information about the target distribution, enabling the model to generalize better in the target domain. The target domain instances $X_t$ are fed into the CTGAN model \cite{xu2019ctgan} to generate synthetic target training data $\tilde{X}_{t}$. It is important to note that the TransCORALNet approach is an unsupervised domain adaptation technique that does not utilize the labels of the target training data. Therefore, the CTGAN approach only utilizes the target domain instances ($X_t$) to generate synthetic target training data. The target domain data contains a mix of discrete and continuous columns. Discrete values can be encoded as one-hot, whereas representing continuous values requires more processing. 

For the sake of simplicity, let us assume that the target domain instances has several continuous columns and only two discrete columns ($D_{1}$ and $D_{2}$). We first select a specific category (Category 1) from $D_{2}$. CTGAN can evenly explore all possible discrete values during the model training process. Next, we choose a row (i.e.\ one instance from $X_t$) with $D_{2}=1$ from the target domain. Then the steps of CTGAN can be summarized as follows. The continuous value will first be normalized by mode-specific normalization \cite{xu2019ctgan}, then it is encoded as one-hot. The selected row with mixed data types is represented as a concatenation of continuous and discrete columns \cite{xu2019ctgan}. CTGAN also introduces a conditional vector \cite{xu2019ctgan} as the representation of all discrete values in the selected row at $D_{1}$ and $D_{2}$, indicating the selected category $D_{2}=1$. The conditional generator receives the conditional vector as input and outputs the conditional distribution of the selected row at the specific category ($D_{2}=1$). The cross-entropy is used as the generator loss to penalize any deviation. The output generated by the conditional generator is assessed by the critic, which estimates the distance between the learned conditional distribution and the real-data conditional distribution of the selected row. 
%at conditional vector.

\subsubsection{Transformer embeddings}

The source domain inputs $X_{s}$ and synthetic target data $\tilde{X}_{t}$ are 
fed into the Transformer encoders, see the upper right of Fig.\ref{fig1}. In what follows, we explain the process within the upper stream corresponding to the source domain $X_{s}$. Notably, this process is the same within the lower stream of the target domain $\tilde{X}_{t}$. The encoder \cite{vaswani2017attention} consists of multi-head self-attention layers and a feed-forward layer followed by layer normalization and residual connections. In our case, in multi-head self-attention layers, each head calculates attention scores for each feature based on its relationship with all other features. In particular, each head transforms the input data to Queries ($Q$), Keys ($K$), and Values ($V$) by multiplying the input embeddings by learned weight matrices
$W^{Q}, W^{K}, W^{V}$ respectively as follows  \cite{vaswani2017attention}:

\begin{equation}
Q= X_s W^{Q}, K= X_s W^{K}, V= X_s W^{V}, 
\end{equation}
where $Q, K, V \in \mathbb{R}^{M\times d_k}$, $d_k =d/h$, $d$ is the dimension of self-attention layer input and output, h is the number of heads.

Next, the attention scores of $j$-th head $A_{j}$ are calculated between the Query and the Key using the scale dot product, shown in Eq. (\ref{eq:2}). These dot products represent the similarity measure between the Query and the Key matrices, indicating how important each Key matrix is to the Query. The attention scores are then normalized using the soft-max function, to generate the attention weights, ensuring that they sum to one and represent the importance or attention scores assigned to each key vector for the given query \cite{vaswani2017attention}. 

\begin{equation}
A_{j}=\operatorname{softmax}\left(\frac{Q K^{T}}{\sqrt{d_k}}\right), A_{j} \in \mathbb{R}^{M\times M}.\label{eq:2}
\end{equation}

The attention outputs of $j$-th head $H_{j}$ are computed by the attention scores and Values (see Eq.(\ref{eq:3})). The outputs of all heads are concatenated and projected, resulting the outputs of the multi-head self-attention layer $H_{multi}$ as shown in Eq. ({\ref{eq:4}) \cite{vaswani2017attention}:

\begin{equation}
H_{j}= A_{j} V ,  H_{j}\in \mathbb{R}^{M\times d_k},\label{eq:3}
\end{equation}

\begin{equation}
H_{multi}=\left(\mathop{\|}\limits_{j=1}^{h} H_{j}\right)W^{o}, H_{multi} \in \mathbb{R}^{M\times d}, \label{eq:4}
\end{equation}
where $W^O \in \mathbb{R}^{h d_k\times{d}}$ is the projection matrix.

In addition to a multi-head attention layer, the encoder block encompasses two fully connected feed-forward layers. Within each multi-head attention layer and the feed-forward layer, the output are processed through residual connections \cite{szegedy2017inception} and layer normalization \cite{ba2016layer}. In a residual connection \cite{szegedy2017inception}, the output of a layer is added to the input of that same layer. Residual connections avoid the problem of exploding and vanishing gradients, improving the ability of the network to learn features. Layer normalization \cite{ba2016layer} works by normalizing the activations of a layer for each sample. It helps stabilize training, reduce sensitivity to initialization, and improve the generalization of deep neural networks.

Next, the output of the transformer encoder block of each stream is fed to five fully connected layers with a ReLU activation function that learn a final representation (feature) of the data.

\subsubsection{Loss function}

Similar to \cite{sun2016deep}, in order for the final deep features to be both discriminative and invariant to the difference between source and target domains, a combination of CORAL loss \cite{sun2016deep} and a classification loss is employed. The CORAL loss \cite{sun2016deep} is used to adjust the distance between second-order statistics (covariance) of the source and target distributions.
%After that, the overall loss function minimizes the Deep CORAL loss and the weighted cross-entropy loss, aiming to learn representations that are both discriminative and minimize the distance between the source and target domains. 
%The accuracy of default borrower prediction directly impacts the profitability of financial institutions. 
In our data, the credit assessment of the supply chain faces an imbalanced class problem, with defaulting samples representing only approximately eleven percent of the total. Therefore, we use the weighted binary cross-entropy loss \cite{ho2019real} as the classification loss which addresses this issue by assigning different weights to each class depending on its relative frequency in the dataset. These weights give more importance to the minority classes and allow the model to learn more effective representations of them. The total loss function is defined as follows:
\begin{equation}
\ell_{total} =\ell_{weighted} +  \lambda \ell_{coral},
\end{equation}
here $\ell_{weighted}$ is the weighted binary cross entropy loss, $\ell_{coral}$ is the CORAL loss. $\lambda$ is the regularization parameter that controls how much importance should be given to the CORAL loss, trading off the adaptation with classification accuracy on the source domain. The weighted binary cross entropy loss is defined as follows \cite{ho2019real}: 
\begin{equation}
\ell_{weighted}=-w\cdot y_{s} \cdot \log \hat{y} -\left(1-w\right)\left(1-y_{s}\right) \cdot \log \left(1-\hat{y} \right),
\end{equation}
where  $w$ represents the weight of the loss function, $y_{s}$ and $\hat{y} $ are the true and predicted label of the sample.

The CORAL loss \cite{sun2016deep} is formulated as:
\begin{equation}
\ell_{coral}=\frac{1}{4 d^{2}}\left\|C_{s}-C_{t}\right\|_{F}^{2},
\end{equation}

\begin{equation}
C_{s}=\frac{1}{M-1}\left(X_{s}^{\top} X_{s}-\frac{1}{M}\left(\mathbf{1}^{\top} X_{s}\right)^{\top}\left(\mathbf{1}^{\top} X_{s}\right)\right),
\end{equation}

\begin{equation}
C_{t}=\frac{1}{M-1}\left(\tilde{X}_{t}^{\top}\tilde{X}_{t}-\frac{1}{M}\left(\mathbf{1}^{\top} \tilde{X}_{t}\right)^{\top}\left(\mathbf{1}^{\top}  \tilde{X}_{t}\right)\right),
\end{equation}
where $d$ represent the data dimension, $\|\cdot\|_{F}^{2}$ denotes the squared Frobenius norm, $C_{s}$ and $C_{t}$ are the covariance matrices of two domains respectively, $M$ is the number of source and synthetic target training data respectively, a vector of all ones with size M is denoted by $\mathbf{1}$.

\subsection{Interpreting the predictions}
LIME \cite{ribeiro2016lime} is a technique designed to explain the predictions of complex black-box machine learning models by approximating them with simpler, interpretable models, such as linear regression or decision trees. Its primary objective is to generate local explanations for the instance $x_s$ and its black-box model predictions $\hat{y}$ by training an interpretable linear model. LIME fits an interpretable local model to the instance $x_s$ and its similar neighbors. The results can be shown by analysing this local model to extract feature importance scores or coefficients, which provide insight into which features had the most influence on the prediction for the specific instance.

Given our proposed black-box model, denoted as $f$, which takes a selected instance $x_s$ needing explanation and produces a predicted output $\hat{y}$. Then define an explainable model $ g \in G$, where $G$ is a class of potentially interpretable models, 
such as linear models and decision trees. Since not all approximators $g$ may be sufficiently simple to be interpretable, the complexity of $g$ is measured by $\Omega(g)$. Furthermore, generate a dataset of perturbed instances with associated labels for the selected instance $x_s$ \cite{ribeiro2016lime}. The perturbed instances are sampled from a simple distribution, such as a Gaussian distribution. These perturbed instances represent the local neighborhood around the sample that need to be explained. Let $\Pi_{x_s}$ represents the proximity measure between the instance $x_s$ and perturbed instances around $x_s$, thereby defining the locality around $x_s$. Finally, let $\mathcal{L}(f, g, \Pi_{x_s})$ be a measure of how unfaithful $g$ approximates $f$ within the locality defined by $\Pi_{x_s}$. In order to ensure both interpretability and local fidelity, LIME minimizes $\mathcal{L}(f, g, \Pi_{x_s})$ while ensuring that $\Omega(g)$ is sufficiently low to enable interpretability. The explanation $\xi(x_s)$ produced by LIME is as follows \cite{ribeiro2016lime} :

\begin{equation}
\xi(x_s)=\underset{g \in G}{\operatorname{argmin}} \mathcal{L}\left(f, g, \Pi_{x_s}\right)+\Omega(g).
\end{equation}

\section{Experiment}\label{section:sec4}
\subsection{Datasets}
The supply chain credit data are collected from a Chinese commercial bank from 2015 to 2021. The source domain contains 82,227 samples from 453 distinct supply chain circles. The average number of samples of each supply chain circle in the source domain is 182 (maximum 15,336 and minimum 3). The target domain comprises 5,632 samples from 80 supply chain circles, entirely different from the source domain. The average number of samples in each supply chain circle in the target domain is 81 (maximum 340 and minimum 11). The classes of two domain samples are binary: positive for defaulting and negative for non-defaulting. The source and target domain defaulting rates are 13\% and 11\%, respectively. 

The samples in two domains are described by the same 21 features (see Table \ref{tbl2}). These features can be divided into repayment willingness and capability factors. Repayment willingness refers to a borrower's intention or motivation to repay a loan. Financial institutions assess borrowers' willingness to repay by evaluating their credit history, financial behavior, and other factors that may suggest a borrower's likelihood to repay a loan. Repayment capability refers to a borrower's ability to repay a loan. Financial institutions evaluate borrowers' repayability by examining their income, expenses, and other financial obligations. 

\begin{table}[width=1.0\linewidth,cols=4,pos=h]
\caption{Features description}\label{tbl2}
\begin{tabular*}{\tblwidth}{@{} LLLL@{} }
\toprule
  & \textbf{feature name} & \textbf{data type}\\
\midrule
\multirow{11}*{Willing} & business scale & category  \\
 & business nature& category  \\
 & company type & category  \\
 & status & category  \\
 & technological enterprise & category  \\
 & government platform finance & category  \\
 & prohibited industry & category  \\
 & listed company & category  \\
 & small and micro enterprises  & category  \\
 & platform type & category  \\
 & years relationship with bank & number  \\
\midrule
\multirow{10}*{Capability} & bank early warning & category  \\
 & guarantee type & category  \\
 & revolving credit facility & category  \\
 & repayment method & category  \\
 & rate adjustment frequency & category  \\
 & credit rating & category  \\
 & deposit balance & number  \\
 & average daily deposit balance & number  \\
 & credit balance & number  \\
 & average daily credit balance & number  \\
\bottomrule
\end{tabular*}
\end{table}

\subsection{Experiment setup}
We split the source domain samples into 80\% source training and 20\% source validation dataset. At the same time, an equivalent number of synthetic target training data is generated using the instances of the target domain by the CTGAN approach. To assess our model performance under varying target domain data distributions, we employ the Kullback-Leibler (KL) Divergence \cite{kullback1951information} between the probability distribution of the target domain and that of the source domain. KL Divergence is one of the most commonly used methods for testing domain shift in machine learning. It measures the difference between two probability distributions, source domain instances $X_s$ and target domain instances $X_t$, by calculating the amount of information lost when approximating $X_s$ with $X_t$. 

The KL divergence value indicates dissimilarity between the source and target domain distributions, with a higher value indicating a greater degree of domain shift. Therefore, we calculated the KL divergence between each target domain circle and the entire source domain. Following this, we arranged the KL divergence values of 80 circles in ascending order. To test whether our model performs consistently well as the domain shift increases, we divided target domain into six groups (see Fig. \ref{fig2}) based on their KL divergence values: 80, 60, 40, 30, 20, and 10. The first group consists of all 80 supply chain circles, whereas the sixth group consists of 10 circles with the highest KL divergence values. The average KL divergence of these groups gradually increases from the first to the sixth group, where the sixth group is the furthest from the source domain. We evaluate our model across these six distinct groups to assess its performance under increasing levels of domain shift.

\begin{figure}[htpb]
    \centering
    \includegraphics[width=8cm]{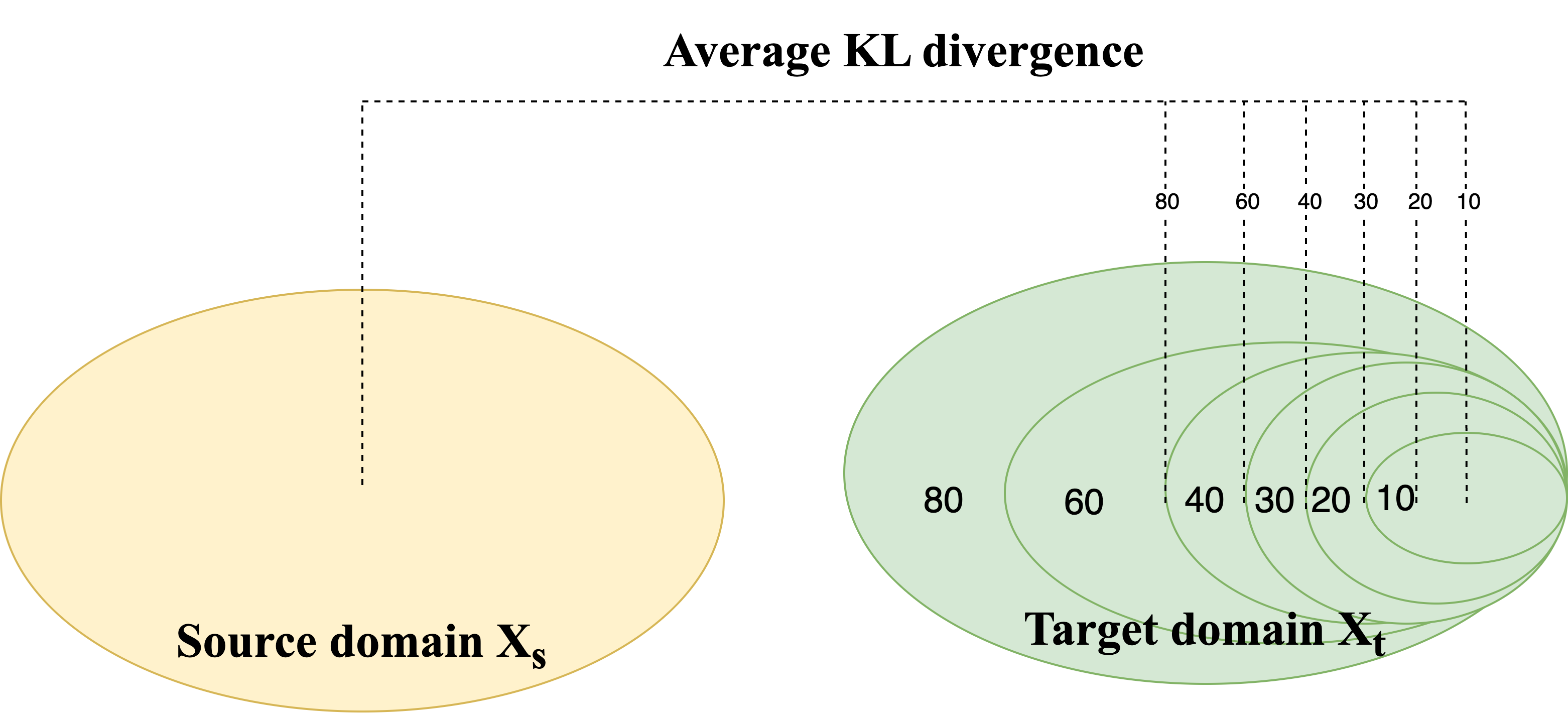}
    \caption{Compute the Kullback-Leibler (KL) divergence for 80 supply chain circles of the target domain relative to the entire source domain, and divide the target domain into six sequential groups based on the average KL divergence within each group, ranging from close to distance.}
    \label{fig2}
\end{figure}

\subsection{Benchmark}
We compared our proposed method with nine benchmark models: 
\begin{itemize} 
\item Four conventional machine learning approaches: logistic regression (LG) \cite{hosmer2013applied}, 
CART decision tree (CART) \cite{lewis2000introduction}, 
random forest 
(RF) \cite{biau2016random}, 
and supporting vector machine (SVM) \cite{hearst1998support}. Note that though these four approaches are essential to verify that conventional machine learning cannot solve the domain drift problem on supply chain credit assessment.
These approaches are trained on the source domain and tested on the target domain, without using weighted loss and synthetic data.

\item Fully connected neural network model (DNN) \cite{schwing2015fully}, this model serves as a \lq no-adaptation\rq, \lq no-weighted-loss\rq and \lq no-attention\rq classifier. It is considered a baseline to compare the effect of domain adaptation and weighted loss.

\item Two discrepancy-based domain adaptation approaches: MMD \cite{tzeng2014deep} and Deep CORAL \cite{sun2016deep}. We use synthetic target data during the training. Besides, a weight was assigned to the cross entropy loss to optimize the imbalanced data problem; 

\item Two adversarial-based domain adaptation approaches: DANN \cite{ganin2016domain} and CADA \cite{long2018conditional}. The synthetic target data were used. Similar to the discrepancy-based domain adaptation approaches, these two approaches are equipped with \lq adaptation\rq and \lq weighted-loss\rq, but do not incorporate an attention mechanism.
\end{itemize}

We substituted the convolutional layers with fully connected layers in the context of four domain adaptation methodologies to adapt the tabular data. Table \ref{tbl3} shows the setting of the baseline model.

\begin{table}[width=1.0\linewidth,cols=6,pos=h]\scriptsize
\caption{Baseline and proposed models settings}\label{tbl3}
\begin{tabular*}{\tblwidth}{@{} LLCCCC@{} }
\toprule
 &  & \textbf{domain}  & \textbf{synthetic}  & \textbf{weighted}  & \textbf{attention}\\
\textbf{types} & \textbf{approach} & \textbf{adaptation} & \textbf{data} & \textbf{loss} & \textbf{mechanism} \\
\midrule
\multirow{4}*{conventional} & LG & & & & \\
& DC & & & & \\
& RF & & & & \\
& SVM & & & & \\
\midrule
\centering
\multirow{1}*{deep learning} & DNN & & & & \\
\midrule
\multirow{2}*{adverisaral DA}& DANN & $\surd $ & $\surd $ & $\surd $ & \\
& CADA & $\surd $ & $\surd $ & $\surd $ & \\
\midrule
\multirow{2}*{discrepancy DA}& MMD & $\surd $ & $\surd $ & $\surd $ & \\
& DeepCORAL & $\surd $ & $\surd $ & $\surd $ & \\
\midrule
\multirow{1}*{proposed}& TransCORALNet& $\surd $ & $\surd $ & $\surd $ & $\surd $ \\
\bottomrule
\end{tabular*}
\end{table}

\begin{table*}[b]\scriptsize
\newcommand{\tabincell}[2]{\begin{tabular}{@{}#1@{}}#2\end{tabular}}
\centering
\caption{The obtained recall and F1 socres of minority class (defaulting) on training, validation and different testing groups using the proposed TransCORALNet and other examined models. }\label{tbl4}
\centering
\begin{tabular}{p{1.6cm} p{1.6cm} p{0.4cm} p{0.4cm} p{0.4cm} p{0.4cm} p{0.4cm} p{0.4cm} p{0.4cm} p{0.55cm} p{0.4cm} p{0.4cm} p{0.4cm} p{0.4cm} p{0.4cm} p{0.4cm} p{0.4cm} p{0.4cm}}\toprule[1pt]
\multirow{3}{*}{\textbf{types}}&\multirow{3}{*}{\textbf{model}} & \multicolumn{8}{c}{\textbf{recall on minority class}} & \multicolumn{8}{c}{\textbf{F1 scores on minority class}}\\ 
 \cmidrule(r){3-10} \cmidrule(r){11-18}
& & \tabincell{c}{train\\-ing}& \tabincell{c}{valida\\-tion\\}& 80& 60& 40& 30& 20& 10& \tabincell{c}{train\\-ing\\}& \tabincell{c}{valida\\-tion\\}& 80& 60& 40& 30& 20& 10\\
\toprule[1pt]
\text{conventional}&LG&0.40 &0.33 &0.13 &0.11 &0.11 &0.11 &0.11 &0.08 &0.35 &0.35 &0.22 &0.22 &0.19 &0.19 &0.19 &0.14 \\
&CART&0.59 &0.59 &0.31 &0.29 &0.25 &0.25 &0.23 &0.12 &0.65 &0.64 &0.41 &0.40 &0.35 &0.36 &0.33 &0.19 \\
&RF&0.71 &0.63 &0.43 &0.41 &0.36 &0.36 &0.35 &0.29 &\textbf{0.77} &0.69 &0.52 &0.53 &0.48 &0.49 &0.48 &0.42 \\
&SVM&0.49 &0.49 &0.32 &0.32 &0.29 &0.29 &0.28 &0.13 &0.57 &0.58 &0.42 &0.43 &0.39 &0.40 &0.39 &0.22 \\
&&&&&&&&&&&&&&&&&\\
\text{deep learning}&DNN&0.69 &0.67 &0.59 &0.55 &0.51 &0.51 &0.41 &0.35 &0.72 &0.70 &0.61 &0.61 &0.58 &0.58 &0.51 &0.34 \\
&&&&&&&&&&&&&&&&&\\
\text{adverisaral DA}&CADA&0.89 &\textbf{0.89} &0.76 &0.74 &0.71 &0.70 &0.63 &0.54 &0.71 &0.66 &0.58 &0.58 &0.55 &0.53 &0.51 &0.38 \\
&DANN&0.87 &0.82 &\textbf{0.80} &0.74 &0.74 &0.74 &0.68 &0.62 &0.70 &0.67 &0.59 &0.59 &0.55 &0.54 &0.53 &0.44 \\
&&&&&&&&&&&&&&&&&\\
\text{discrepancy DA}&MMD&0.89 &0.87 &0.74 &0.74 &0.70 &0.70 &0.62 &0.52 &0.73 &\textbf{0.72} &0.67 &0.67 &0.62 &0.62 &0.59 &0.49 \\
&DeepCORAL&0.89 &0.88 &0.76 &0.76 &0.72 &0.72 &0.68 &0.62 &0.73 &\textbf{0.72} &\textbf{0.69} &0.68 &0.65 &0.65 &0.64 &0.57 \\
&&&&&&&&&&&&&&&&&\\
\text{proposed}&TransCORALNet&\textbf{0.90} &\textbf{0.89} &0.78 &\textbf{0.78} &\textbf{0.77} &\textbf{0.77} &\textbf{0.72} &\textbf{0.67} &0.73 &0.71 &\textbf{0.69} &\textbf{0.69} &\textbf{0.68} &\textbf{0.68} &\textbf{0.68} &\textbf{0.62}\\
\midrule[1pt]
\end{tabular}
\label{table:stu}
\end{table*}

\begin{figure}[htpb]
    \centering
    \includegraphics[width=8cm]{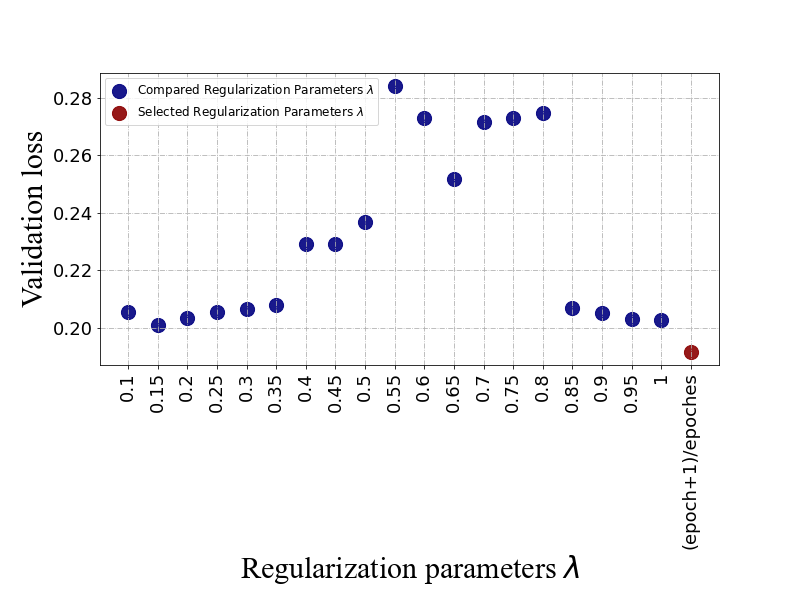}
    \caption{The loss on validation dataset corresponding to different regularization parameters $\lambda$}
    \label{fig3}
\end{figure}

\subsection{Training}
In the employed conventional machine learning approaches, a 10-fold cross-validation is implemented for the model selection. The other tested approaches (DNN, DANN, CADA, MMD and DeepCORAL), as well as our proposed model, are trained for maximum 250 epochs. We employed an early stopping criterion which stopped the training process when the validation loss did not decrease in the last 15 epochs. The training process employed the stochastic gradient descent (SGD) optimizer with default settings. The training procedure is performed iteratively using a batch size of 256. Additionally, we used a learning rate scheduling that reduced the learning rate over time by using exponential decay. We set the initial learning rate to 0.1. The optimal hyperparameters, specifically the weight assign to cross-entropy loss, are empirically found to be 0.25 for non-defaulting samples and 0.75 for defaulting samples. Additionally, the optimal number of heads for multi-head self-attention layers is empirically found to be seven.

To determine the optimal value for the regularization parameter $\lambda$, we adopt a search strategy and investigate 19 different values in the interval [0.1, 1], $\lambda  \in \left \{ 0.1, 0.15, \dots, 1 \right \}$. Besides, we consider a epoch-varying $\lambda$ that increased gradually from 0 to 1 with the progression of training, defined as $\lambda = (epoch + 1)/epochs$, where {\lq epoch\rq} represents the current epoch, {\lq epochs\rq} is the total epochs. We compare the validation data loss under these settings, and the influence of the regularization parameter $\lambda$ is displayed in Fig. \ref{fig3}. Finally, we set a epoch-varying $\lambda$ based on these observations.

\begin{figure*}[ht]
  \centering
  \subfloat[] 
  {\includegraphics[width=8cm]{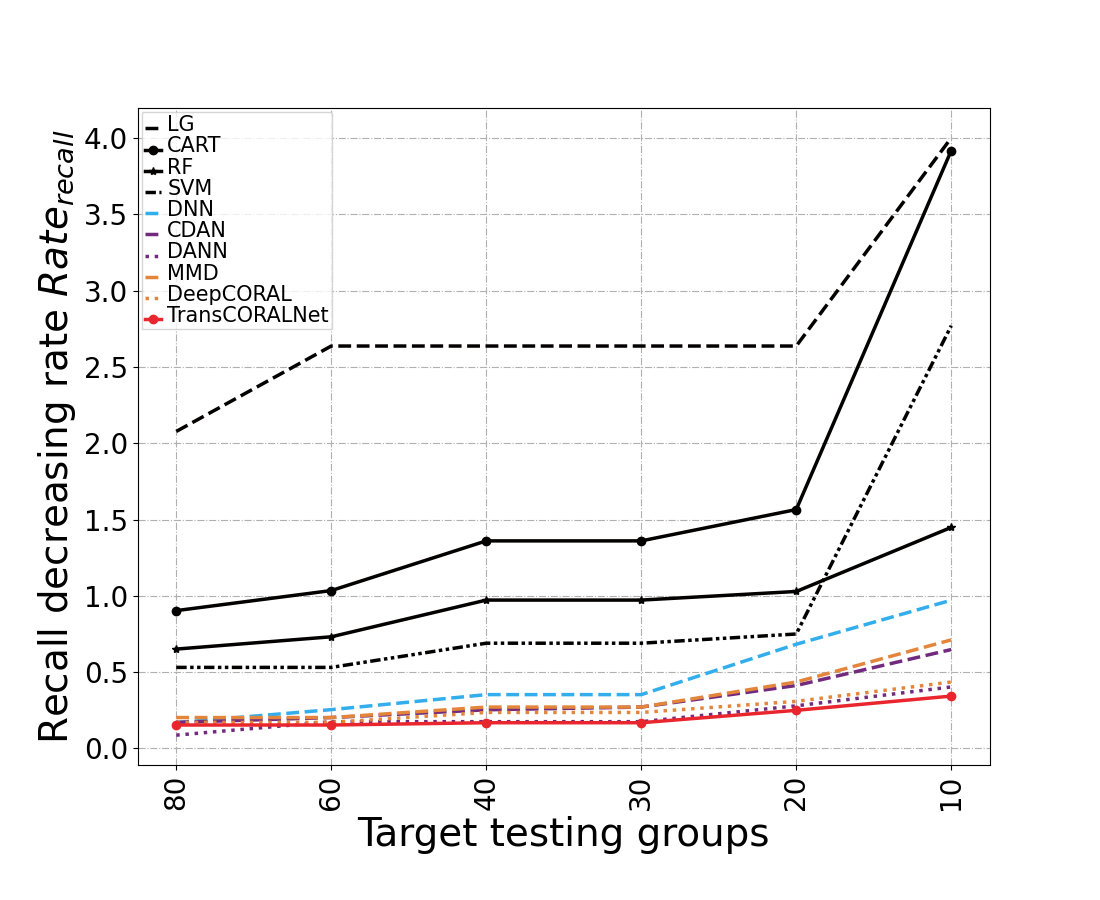}\label{fig4:f1}}
  \hfill
  \subfloat[]
  {\includegraphics[width=8cm]{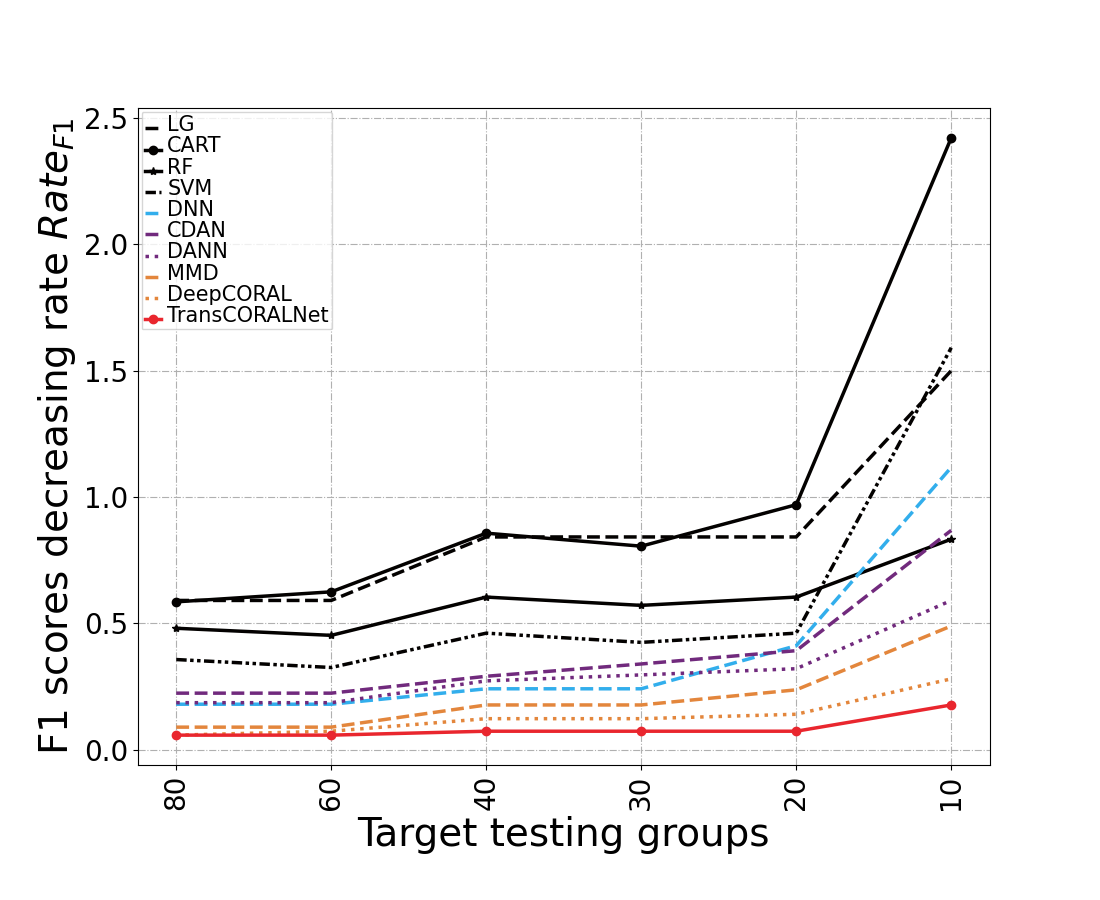}\label{fig4:f2}}
  \caption{Comparison of recall and F1 scores decreasing rate of proposed model and benchmark modes on different target test groups. (a) Recall decreasing rate. (b) F1 scores decreasing rate. }\label{fig4}
\end{figure*}

\begin{figure}[htpb]
\centering
\includegraphics[width=8cm]{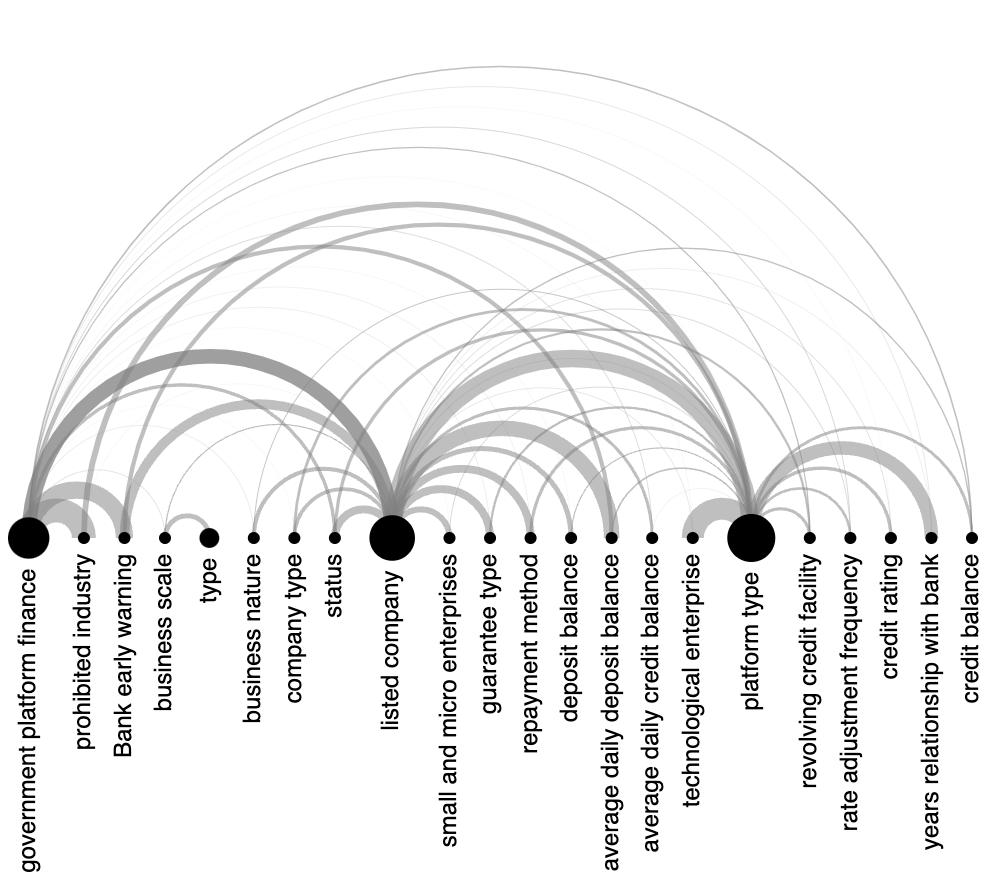}
\caption{The average attention scores of all instances in the source domain. The three biggest nodes represent the three most important features: government platform finance, platform type and listed company. The broader and darker lines indicate the larger attention score.}\label{fig5}
\end{figure}

\begin{figure}[htpb]
  \centering
  \subfloat[]
  {\includegraphics[width=7.6cm]{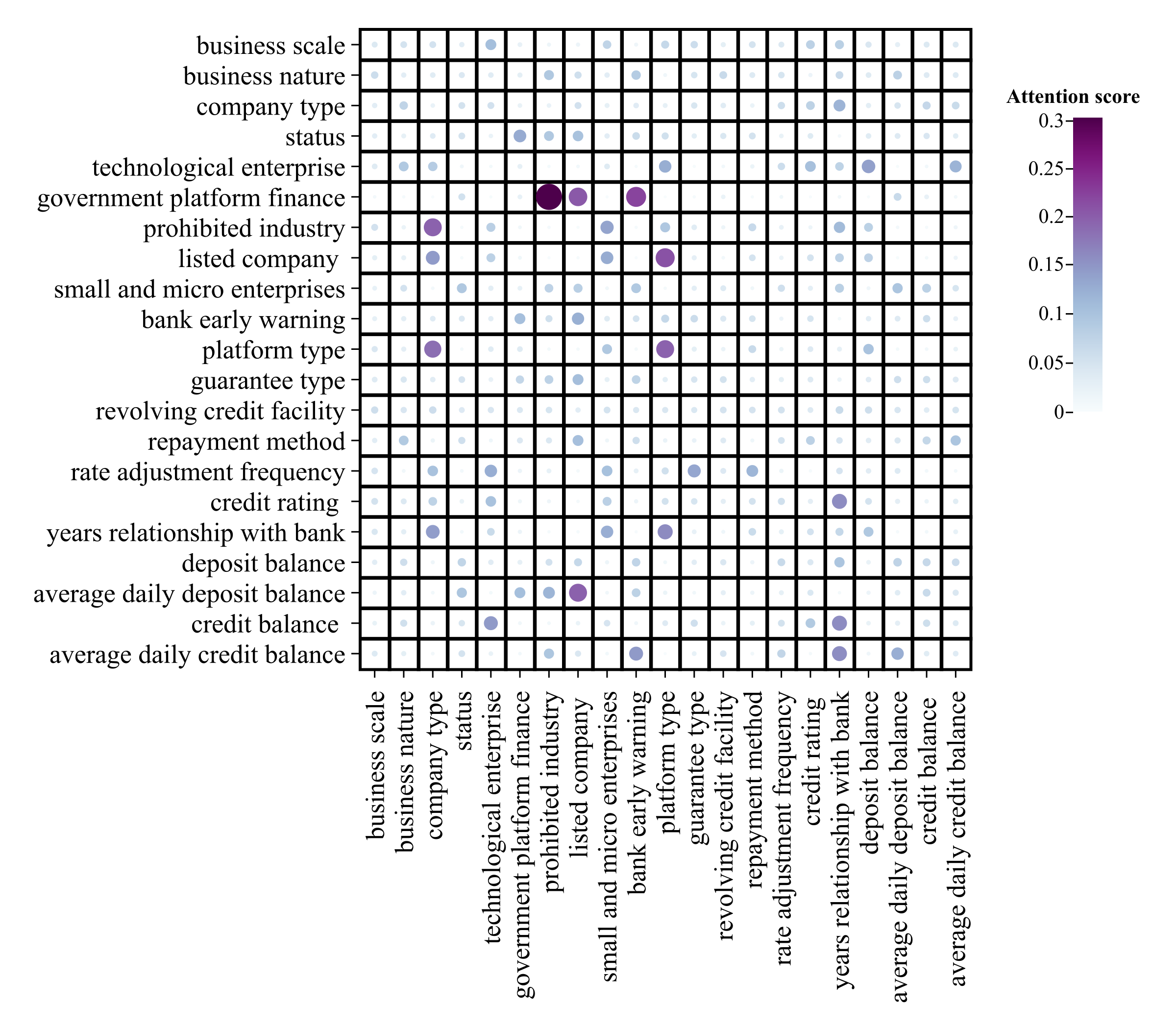}\label{fig6:f1}}
  \hfill
  \subfloat[]{\includegraphics[width=7.6cm]{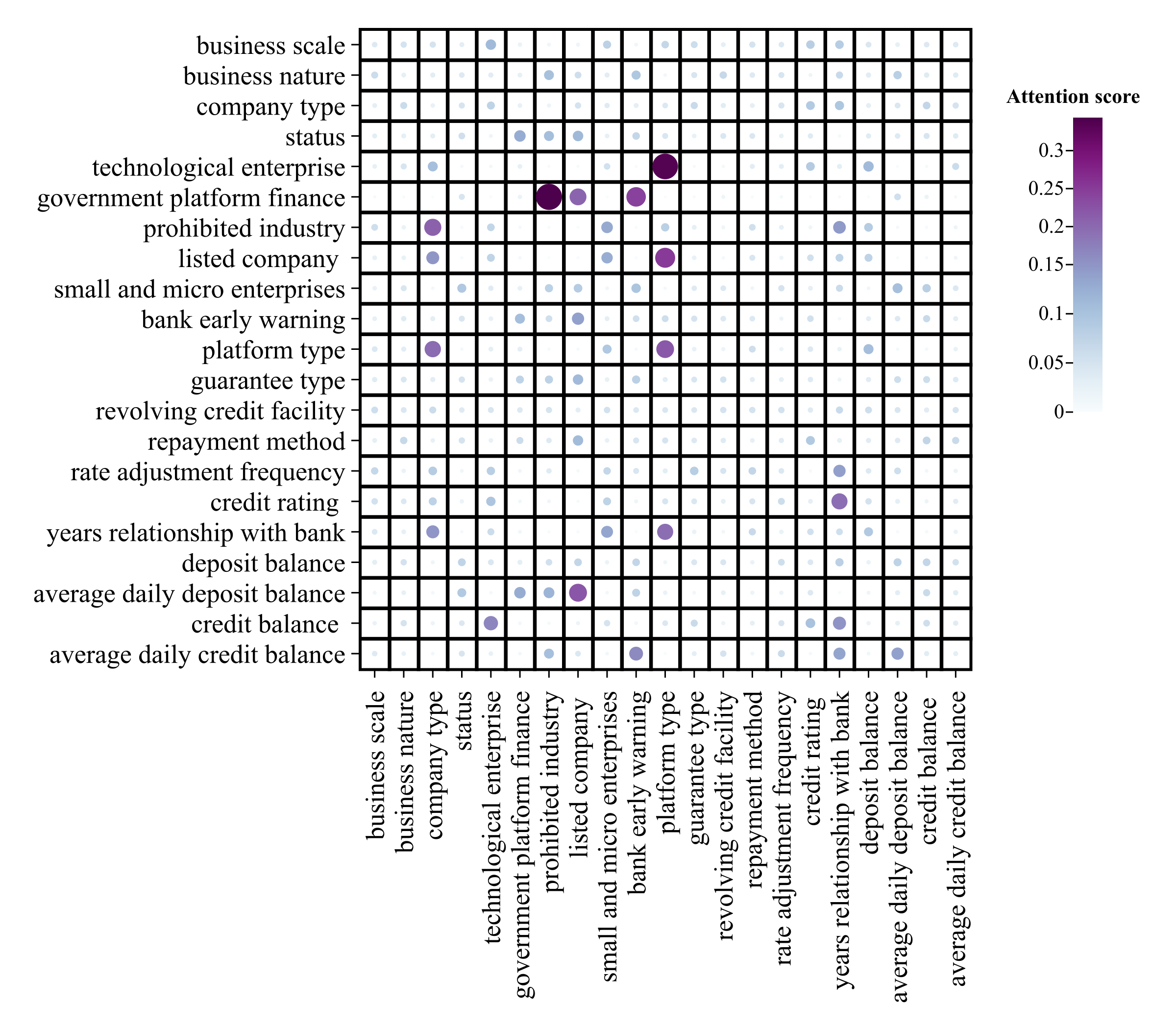}\label{fig6:f2}}
  \hfill
  \subfloat[]{\includegraphics[width=7.6cm]{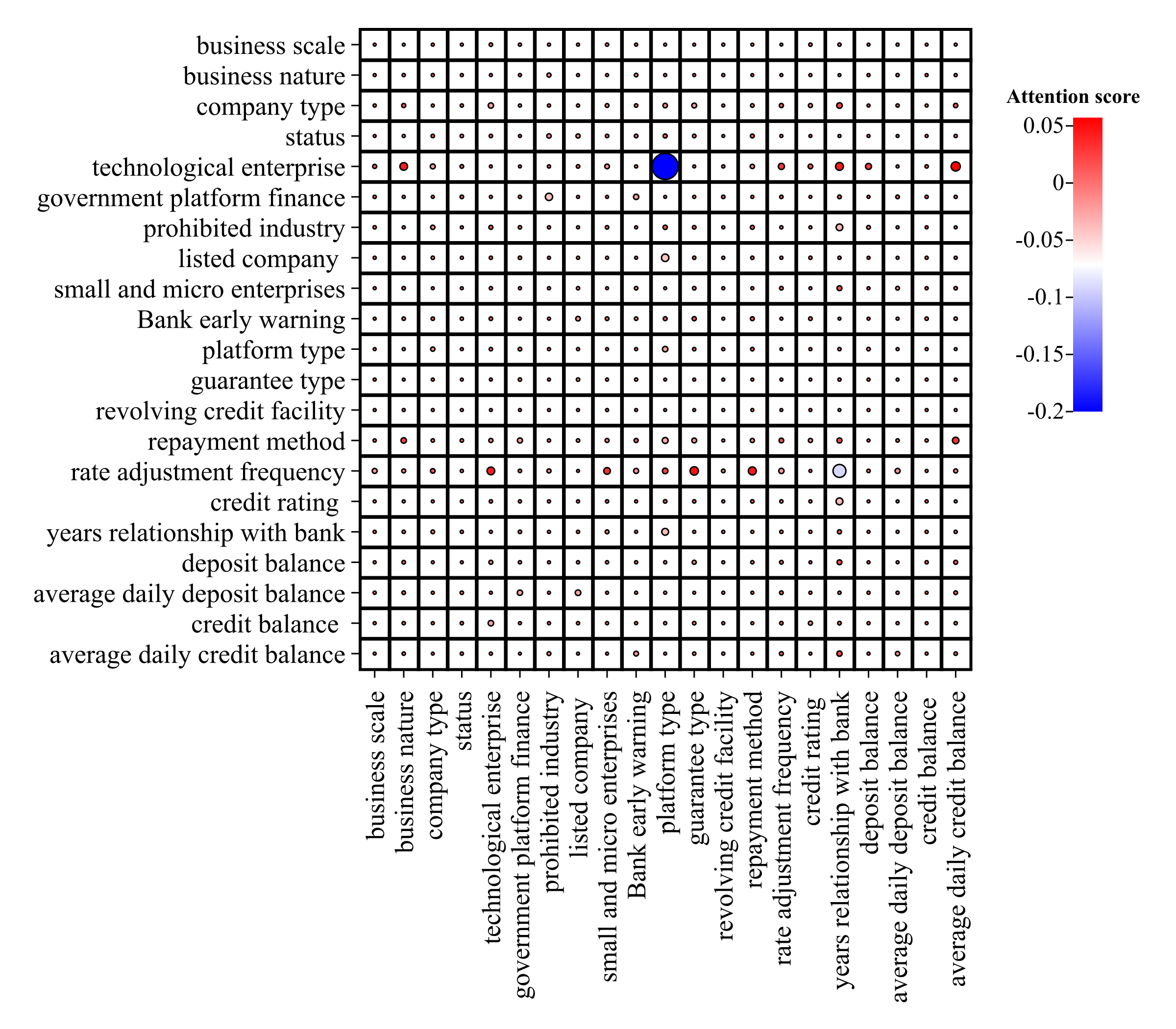}\label{fig6:f3}}
  \caption{(a) The mean attention scores of defaulting samples. (b) The mean attention scores of non-defaulting samples. (c) Attention scores differences, which is the mean attention scores of defaulting samples subtract that of non-defaulting samples.}\label{fig6}
\end{figure}

\subsection{Model evaluation}
Defaulting borrowers can directly lead to a loss of profits for financial institutions. The consequences of false negatives significantly outweigh those of false positives in credit assessment model prediction. Hence, we use the recall on the positive class as our model indicator. On the other hand, mislabeling a regular borrower as defaulting can lead to customer churn. In this case, we use the F1 score as a trade-off solution. The recall and F1 scores are calculated on the positive data. These indicators \cite{goutte2005probabilistic} are formulated as follows:
\begin{equation}
\mathrm{Recall}=\frac{T P}{T P+F N} \times 100 \%,
\end{equation}

\begin{equation}
\mathrm{Precision}=\frac{T P}{T P+F P} \times 100 \%,
\end{equation}

\begin{equation}
\textrm{F}_{1}=2 \times \frac{\textrm { Precision } \times \textrm { Recall }}{\textrm { Precision }+ \textrm { Recall }},
\end{equation}
where TP (prediction=defaulting, true=defaulting), FP (prediction=defaulting, true=non-defaulting) and FN(prediction
=non-defaulting, true=defaulting) represent the number of true positive, false positive and false negative of model prediction respectively.

\section{Results and discussion}\label{section:sec5}

The performance of our proposed TransCORALNet model and 
other tested approaches are depicted in Table \ref{tbl4}. It shows the recall and F1 scores on the minority (defaulting) class of source training, source validation, and target test groups (80, 60, 40, 30, 20, and 10 supply chain circles), respectively. The results indicate that our proposed TransCORALNet model, consistently outperformed the other models in each target testing group, providing the best predictions. It should also be noted that the TransCORALNet achieves the highest recall and F1 scores on the fifth group (the most dissimilar ten groups with source domain). TransCORALNet significantly outperforms the others in addressing the imbalanced classes, cold start, and domain shift problems, highlighting its effectiveness and superiority in solving these challenges. Table \ref{tbl4} also shows that conventional ML approaches (LG, CART, RF, SVM) failed in supply chain credit assessment where there is a domain shift problem. Furthermore, the DNN demonstrated a more competitive performance than the conventional ML approaches but still struggles to address the domain drift problem as the mismatch between source and target domain increased. Four examined domain adaptation approaches (MMD, Deep CORAL, CADA and DANN) demonstrated relatively higher performance than DNN. The results further support that weighted loss and domain adaptation can help the model handle imbalanced data sets and domain shift.

To further evaluate and visualize our proposed model performance on addressing domain shift, we calculate each model's recall and F1 score decreasing rate (see Fig. \ref{fig4}). These decreasing rates can provide insights into how our model demonstrates generalization efficacy from the source to the target data. A lower recall/F1 decreasing rate suggests that the model's performance is relatively stable and generalized in both the source and target domain,  indicating that it is not overly sensitive to variations in the target domain. The decreasing rate of recall $\textrm{Rate}_{recall}$ and $\textrm{Rate}_{F_{1}}$ are computed as follows:
\begin{equation}
\textrm{Rate}_{recall} = \frac{\textrm{Recall}_{training} -\textrm{Recall}_{testing}}{\textrm{Recall}_{testing}},
\end{equation}

\begin{equation}
\textrm{Rate}_{F_{1}} = \frac{\textrm{F}_{1training} -\textrm{F}_{1testing}}{\textrm{F}_{1testing}},
\end{equation}
where $\textrm{Recall}_{training}$ and $\textrm{F}_{1training}$ represent the recall and F1 socore on source training data, $\textrm{Recall}_{testing}$ and $\textrm{F}_{1testing}$ are the recall and F1 score on different target testing groups respectively.

Fig. \ref{fig4} compares the recall/F1 score decreasing rate, representing models' ability to handle domain shift and imbalanced class problems. When the data distributions differ, our proposed model shows the most flattened trend in recall and F1 decreasing rate from 80 to 10 group. In particular, it is noteworthy that, in contrast to the substantial declines observed in the performance of all other models under comparison at the fifth group (10 group), our proposed model exhibited remarkable stability. These results signify the robust performance of TransCORALNet, highlighting its potential as an effective model for supply chain credit assessment.

\begin{figure*}[htpb]
  \centering
  \subfloat[]
  {\includegraphics[width=8.5cm]{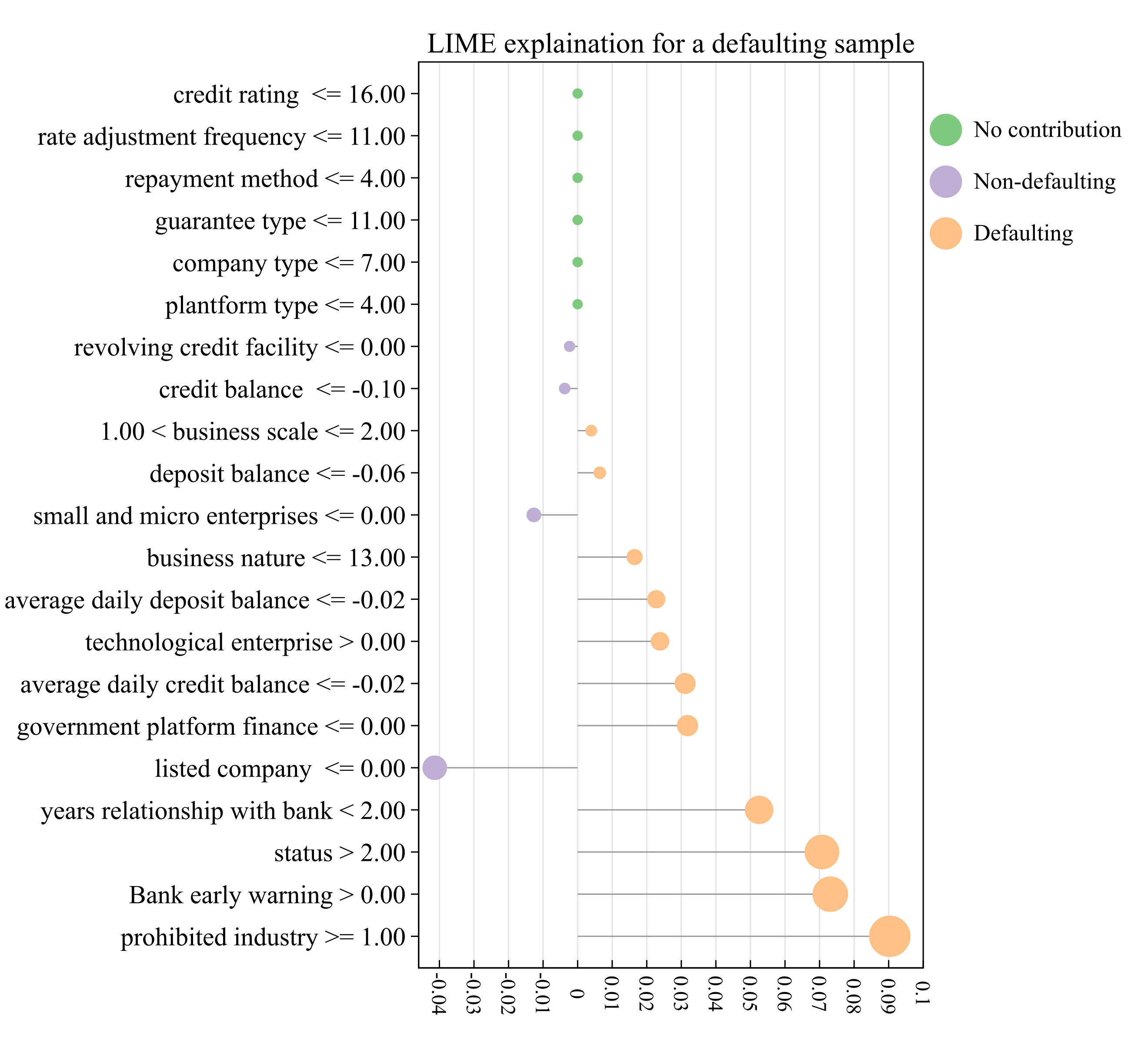}\label{fig7:f1}}
  \hfill
  \subfloat[]{\includegraphics[width=8.5cm]{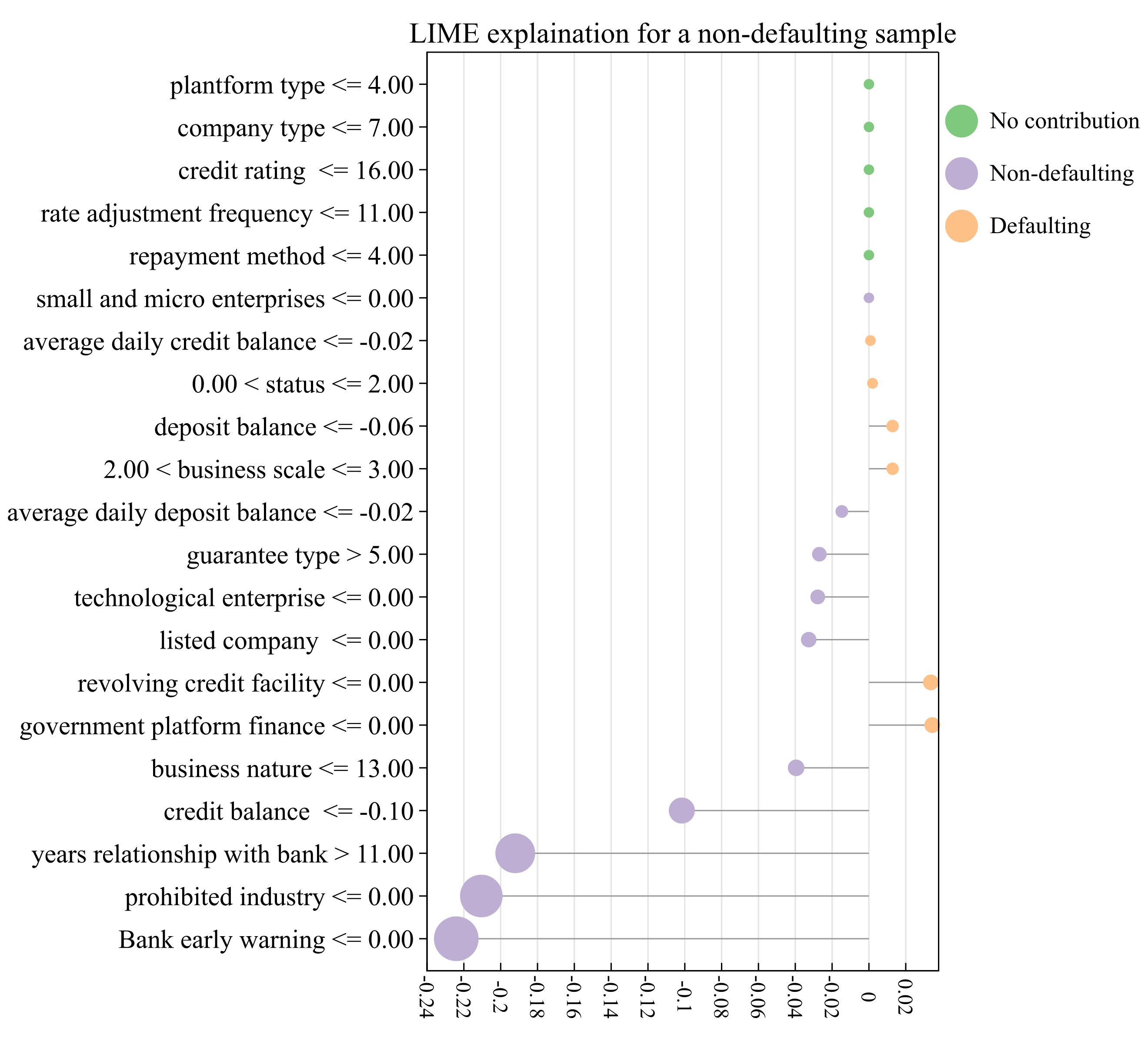}\label{fig7:f2}}
  \caption{Two examples of LIME explanation results. The bars’ length and cricle's wideth highlight the contribution of each variable. (a) Predicted defaulting borrower. (b) Predicted non-defaulting borrower. }\label{fig7}
\end{figure*}

\subsection{Interpretation of attention scores}

Fig. \ref{fig5} presents the average attention score of all source domain instances, assessing the critical features that impact the prediction. This visualization provides insights into the relative importance and relationships of these features. It illustrates the three most crucial features and their interconnectedness with the remaining features.

The {\lq government platform finance\rq}, {\lq platform type\rq} and {\lq listed company\rq} hold significant importance in supply chain credit. Additionally, the feature {\lq government platform finance\rq} shows a substantial correlation with both {\lq prohibited industry\rq} and {\lq bank early warning\rq}. Similarly, the feature {\lq platform type\rq} demonstrates a high degree of relevance assigned to {\lq technological enterprise\rq}. These three features collectively contribute to borrowers' repayment willingness. The findings emphasize that borrower willingness to pay plays a crucial role in predicting defaulting, outweighing other variables commonly considered in defaulting risk assessment. Furthermore, the results demonstrate that the concurrent presence of these particular feature combinations significantly enhances the overall predictive capability of the model.

To identify the essential features contributing to supply chain borrower default, we calculated the average attention scores for both defaulting and non-defaulting samples separately. Fig. \ref{fig6:f1} and \ref{fig6:f2} illustrate the feature attention scores for pairs of defaulting and non-defaulting samples. In the case of defaulting borrowers, the features {\lq government platform finance\rq} exhibit a strong connection with {\lq prohibited industry\rq}, {\lq listed company\rq} and {\lq bank early warning\rq}. Additionally, the {\lq average daily deposit balance\rq} has a higher attention score than the {\lq listed company\rq}. For non-defaulting borrowers, {\lq technological enterprise\rq} and {\lq platform type\rq} show a robust relationship.

To provide a more comprehensive understanding and a clear visualization of the impact of pairwise feature importance on defaulting behavior, we subtract the average attention score of defaulting from that of non-defaulting borrowers. Fig. \ref{fig6:f3} visualizes this new matrix. A higher value signifies a strong correlation between two features, thereby increasing the likelihood of borrower defaulting. Conversely, a lower numerical value suggests that the borrower is expected to be non-defaulting.

When borrowers exhibit a high attention score between {\lq technological enterprise\rq} with {\lq average daily credit balance\rq} and {\lq years relationship with the bank\rq}, {\lq rate adjustment frequency\rq} with {\lq guarantee type\rq} and {\lq repayment method\rq}, they are more likely to default. In contrast, when there is a strong correlation between {\lq technological enterprise\rq} with {\lq platform type\rq}, as well as {\lq rate adjustment frequency\rq} with {\lq years relationship with bank\rq}, the sample has a high probability of being non-defaulting.

\subsection{Model explanation}
Two examples of the LIME explanation results are shown in Fig. \ref{fig7}. The diagrams illustrate how features influence the model prediction and which features indicate a defaulting/non-defaulting classification. The bar diagrams on the right (orange circles) support the {\lq defaulting\rq}, whereas those on the left (purple circles) contradict it. The values of zero (green) indicate {\lq no contribution\rq}. Fig. \ref{fig7:f1} depicts a case predicting a defaulting borrower. Features that increase the likelihood of classification as defaulting, ordered by weights, are {\lq prohibited industry\rq}, {\lq bank early warning\rq}, {\lq status\rq} and {\lq years relationship with bank\rq}. In contrast, the {\lq listed company\rq} decreases the possibility of being categorized as a defaulter. Fig. \ref{fig7:f2} explains why a borrower is more likely to be non-defaulting. Financial institutions can consult the model interpreter's results to provide borrowers with appropriate justifications when declining credit applications.

\section{Conclusion and future work}\label{section:sec6}
In this paper, a two-stream transformer CORAL network (TransCORALNet) for supply chain credit assessment in cold start scenarios is introduced. Our experimental results demonstrate that the proposed model outperforms several state-of-the-art domain adaptation methods for the tested dataset. The self-attention mechanism of TransCORALNet can capture the relationship between the features, which provides insightful explanations on defaulting behaviors of supply chain credit borrowers. Furthermore, we adopt a explanation technique (LIME) to generate explanations focusing on a specific prediction to ensure that financial institutions' decisions are made for the correct reasons. An exciting and important direction for future research is to investigate the effectiveness of TransCORALNet in other applications with similar cold start challenges.

\section{Bibliography}

%\appendix
%\section{My Appendix}
%Appendix sections are coded under %\verb+\appendix+.

%\verb+\printcredits+ command is used after appendix sections to list 
%author credit taxonomy contribution roles tagged using \verb+\credit+ 
%in frontmatter.


\bibliographystyle{unsrt}

\bibliography{transcoral}

\end{document}